\documentclass[preprint,12pt]{elsarticle}

\usepackage[utf8]{inputenc}
\usepackage[T1]{fontenc}
\usepackage{lmodern}
\usepackage{newunicodechar}
\newunicodechar{ℝ}{\ensuremath{\mathbb{R}}}
\newunicodechar{ℕ}{\ensuremath{\mathbb{N}}}
\newunicodechar{→}{\ensuremath{\to}}
\newunicodechar{≤}{\ensuremath{\leq}}
\newunicodechar{∀}{\ensuremath{\forall}}
\newunicodechar{∧}{\ensuremath{\wedge}}
\newunicodechar{δ}{\ensuremath{\delta}}
\newunicodechar{α}{\ensuremath{\alpha}}
\newunicodechar{⟨}{\ensuremath{\langle}}
\newunicodechar{⟩}{\ensuremath{\rangle}}
\newunicodechar{ρ}{\ensuremath{\rho}}
\newunicodechar{∈}{\ensuremath{\in}}
\usepackage{amsmath,amssymb,amsthm}
\usepackage{graphicx}
\usepackage{booktabs}
\usepackage{hyperref}
\usepackage{xcolor}
\usepackage{algorithm}
\usepackage{algpseudocode}
\usepackage{listings}
\usepackage{caption}
\usepackage{subcaption}
\usepackage{multirow}
\usepackage{enumitem}

\lstdefinelanguage{Lean4}{
  morekeywords={theorem,by,calc,rw,exact,have,intro,apply,sorry,
    variable,section,end,import,open,noncomputable,let,simp,linarith,nlinarith,forall,constructor},
  morecomment=[l]{--},
  morecomment=[s]{/-}{-/},
  morestring=[b]",
  sensitive=true,
  literate={→}{{$\to$}}1 {ℝ}{{$\mathbb{R}$}}1 {ℕ}{{$\mathbb{N}$}}1
           {≤}{{$\leq$}}1 {∀}{{$\forall$}}1 {∧}{{$\wedge$}}1
           {δ}{{$\delta$}}1 {δ₁}{{$\delta_1$}}2 {δ₂}{{$\delta_2$}}2
           {⟨}{{$\langle$}}1 {⟩}{{$\rangle$}}1
           {α}{{$\alpha$}}1
           {ρ}{{$\rho$}}1
           {∈}{{$\in$}}1,
}
\journal{Neural Networks}

\begin{document}

\begin{frontmatter}

\title{Structural Sensitivity in Compressed Transformers:\\Relative Error Propagation and Layer Removal}

\author[1,2]{Abhinaba Basu\corref{cor1}}
\ead{mail@abhinaba.com}
\author[3]{Kumkum Basu}
\author[4,5]{Koushik Deb}

\cortext[cor1]{Corresponding author}

\address[1]{Indian Institute of Information Technology Allahabad, India}
\address[2]{National Institute of Electronics and Information Technology (NIELIT), India}
\address[3]{Indian Institute of Technology Patna, India}
\address[4]{Indian Institute of Information Technology Kalyani, India}
\address[5]{Accenture, India}

\begin{abstract}
Compressing transformer weights makes large language models cheaper to deploy. But each layer's compression introduces an error. These errors accumulate as the signal passes through later layers, and how they accumulate is not well understood. We measure this directly: at each layer, we take the ratio of output to input error, calling it $\rho$. A value below one means the layer absorbs the error; above one means it grows. Computing $\rho$ on six transformers (117M to 8B parameters) yields three findings. (i)~Errors at layer $t$ scale downstream by the product of later $\rho$ values, predicting representation drift (Spearman $r = -0.44$, $p < 10^{-4}$). This explains why compressing early layers hurts more than late ones, and why depth-decreasing sparsity schedules outperform uniform ones. Across architecture families, however, model width and redundancy matter more than $\rho$ alone. (ii)~Within a layer, naive pruning shows a ${\sim}600\times$ spread in component sensitivity. Activation-aware pruning (Wanda) shrinks this to $3\text{--}7\times$; the ranking reverses across architectures, so fixed importance scores do not transfer. (iii)~For depth pruning, ranking layers by how far $\rho$ is from one takes two forward passes. It beats ShortGPT's Block Influence with $1.6\times$ lower perplexity at eight layers removed, and physical deletion delivers $1.22\times$ wall-clock speed-up. A blend of the two criteria does best (perplexity 14.2, 60.0\% downstream accuracy on LLaMA-2-7B). Twelve Lean~4 norm inequalities provide machine-checked per-matrix error bounds. The contraction profile thus gives a training-free instrument for two decisions: where to compress within layers, and which to remove.
\end{abstract}


\end{frontmatter}

\section{Introduction}

Large language model inference is increasingly bottlenecked by memory bandwidth rather than compute~\cite{pope2023efficiently,shazeer2019fast}.
This has motivated a rich literature on model compression (pruning~\cite{frantar2023sparsegpt,sun2024wanda}, quantization~\cite{frantar2023gptq,dettmers2022gpt3}, and low-rank factorization~\cite{hsu2022language}) that achieves high sparsity with minimal quality loss on standard benchmarks.
However, a fundamental question remains underexplored: \emph{how do compression errors propagate through transformer layers, and what structural properties of the architecture govern this propagation?}

Understanding error propagation is critical for several reasons.
First, practitioners need to know which components are safe to compress and which are not, without exhaustive empirical sweeps on every new model.
Second, verified per-matrix error bounds enable principled cost--quality trade-offs in deployment.
Third, the mechanisms that cause certain components to be catastrophically sensitive to compression reveal fundamental aspects of transformer computation that are of independent scientific interest.

To address these needs, this paper combines \emph{analysis} with a \emph{practical method}. We develop a Lyapunov-style characterization of compression error propagation, validate it across six architectures from 117M to 8B parameters, and derive a layer-removal criterion that complements existing heuristics. The headline result is practical: physical layer deletion guided by the resulting contraction profiles gives $1.22\times$ real inference speedup on LLaMA-2-7B, and among the pruning-based methods we test, only physical deletion translates to wall-clock gains on commodity GPUs without specialized sparse hardware.

\paragraph{Contributions}
We make four contributions:

\begin{enumerate}[leftmargin=*]
\item \textbf{Structural sensitivity mapping and contraction analysis} (\S\ref{sec:sensitivity}--\S\ref{sec:crossarch}).
We map compression sensitivity at per-matrix granularity across six architectures (117M--8B) and measure per-layer relative error contraction ($\rho_{\max} = 0.96$ on GPT-2 Small).
The sensitivity hierarchy spans five orders of magnitude but collapses to $3\text{--}7\times$ under activation-aware pruning, and within-layer reallocation fails on every architecture tested.
Twelve Lean~4 theorems provide verified per-matrix error accounting (\S\ref{sec:framework}, \S\ref{sec:lean}).

\item \textbf{Cross-architecture validation} (\S\ref{sec:crossarch}).
Contraction is necessary but not sufficient for compression tolerance: among similar architectures, higher $\rho_{\max}$ correlates with worse outcomes, but across families, width-dependent redundancy dominates.

\item \textbf{Contraction-guided layer removal} (\S\ref{sec:layerremoval}).
Layers with $\rho \approx 1$ are empirically low-regret (21 of 32 LLaMA-2-7B layers removable with $<0.5$ PPL regret).
$|\rho - 1|$ ranking yields $1.1\text{--}1.7\times$ lower perplexity than the from-end heuristic; ShortGPT's Block Influence (BI; the cosine distance between a layer's input and output)~\cite{men2024shortgpt} retains better downstream accuracy, and the two criteria are complementary.

\item \textbf{Practical speedup validation} (\S\ref{sec:layerremoval}).
Physical layer deletion gives $1.22\times$ real inference speedup on commodity hardware; weight pruning gives none ($0.96\times$).
A blended criterion ($\lambda|\rho{-}1| + (1{-}\lambda)\text{BI}$) dominates both pure strategies at 8 layers removed across three architectures.
\end{enumerate}

\begin{figure*}[t]
\centering
\includegraphics[width=\textwidth]{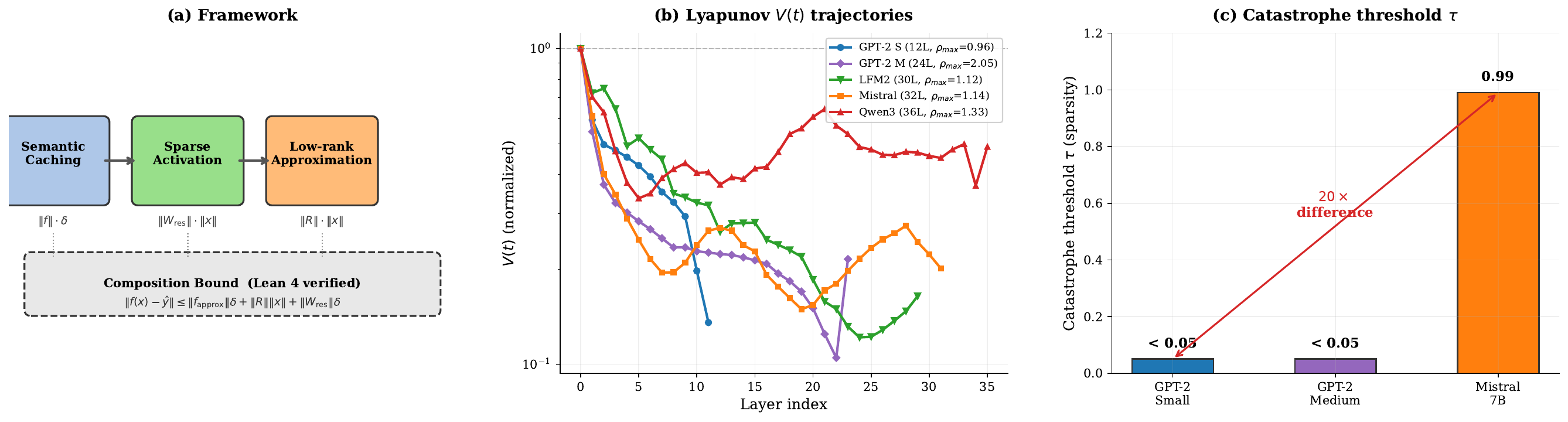}
\caption{Overview. (a)~Three composable approximation stages with per-matrix bounds. (b)~Relative error energy $V(t)$ trajectories across six architectures: GPT-2 Small contracts monotonically ($\rho_{\max} = 0.96$), while GPT-2 Medium's final layer amplifies ($\rho_{23} = 2.05$). LFM2-2.6B (hybrid) and Mistral-7B show moderate amplification; Qwen3-8B oscillates most. (c)~Catastrophe threshold $\tau$ increases with model dimension but is modulated by architecture (Table~\ref{tab:threshold_all}).}
\label{fig:overview}
\end{figure*}

\section{Related Work}
\label{sec:related}

Our work draws on prior research in model compression, formal verification, and stability analysis of deep networks. We summarise the most relevant threads below and position our contributions against each.

\paragraph{Model compression}
Post-training pruning methods such as SparseGPT~\cite{frantar2023sparsegpt} and Wanda~\cite{sun2024wanda} achieve high sparsity with minimal perplexity loss, while GPTQ~\cite{frantar2023gptq} provides effective weight quantization.
These methods are calibrated empirically on held-out data and provide no formal guarantees.
Our framework is complementary: the sparsification and low-rank stages could incorporate these methods while adding provable error bounds.

\paragraph{Formal verification in ML}
Formal verification has been applied to neural network robustness~\cite{katz2017reluplex,cohen2019certified}, training correctness~\cite{selsam2017developing}, and floating-point arithmetic~\cite{boldo2015coquelicot}.
We are not aware of prior work that machine-checks per-matrix error bounds for transformer inference approximation, though the underlying operator-norm inequalities are standard and the bounds cover only the linear components.

\paragraph{Lyapunov-style analysis in deep learning}
Lyapunov methods have been used to analyze training dynamics~\cite{haber2017stable} and residual network stability~\cite{khalil2002nonlinear}, but their application to \emph{inference-time error propagation} in transformers has not been previously explored.
We provide quantitative per-layer contraction measurements ($\rho < 1$ for all GPT-2 Small transitions), going beyond qualitative arguments about normalization.
Our contraction factors are empirical trajectory-level diagnostics, not global stability certificates in the strict dynamical-systems sense.

\paragraph{Norm-growth composition in neural networks}
Lipschitz constant estimation for neural networks has been studied extensively~\cite{virmaux2018lipschitz,fazlyab2019efficient}, but most approaches compute worst-case sublayer bounds that are vacuously large for deep networks---typically $\prod K_i \gg 10^{10}$~\cite{nagarajan2019uniform}. We instead measure $K_i = \|h_{t+1}\|/\|h_t\|$ at the \emph{residual-layer level}, which captures the stabilizing effect of skip connections. This is a trajectory-level norm-growth ratio, not a true Lipschitz constant (which bounds output change under input perturbation); the resulting product ($< 530$ for models up to 7B, CV $< 4\%$) is an empirical structural characteristic rather than a worst-case guarantee, but its non-vacuousness---in contrast to sublayer-level estimates---demonstrates that residual connections fundamentally tighten error-propagation bounds.

\paragraph{Control theory and model reduction}
Balanced truncation~\cite{moore1981principal} and Hankel norm approximation~\cite{glover1984all} are classical model reduction techniques from control theory that provide optimal or near-optimal low-rank approximations with formal error guarantees.
Our application of balanced truncation to transformer weight compression demonstrates that control-theoretic tools can improve upon naive SVD truncation, achieving $2.5\times$ more compressed groups at aggressive compression levels.

\paragraph{Non-uniform compression}
Mixed-precision and non-uniform quantization~\cite{dong2019hawq,yao2022zeroquant} allocate more bits to sensitive layers, and OWL~\cite{yin2023outlier} adapts sparsity based on outlier features.
ATP~\cite{zhuang2025atp} derives layer-wise sparsity ratios from an optimization-theoretic framework, achieving results competitive with SparseGPT.
AlphaPruning~\cite{lu2024alphapruning} (NeurIPS 2024) uses the heavy-tailed spectral properties of weight matrices ($\alpha$ exponent) to allocate non-uniform sparsity, providing a complementary theoretical lens.
Wanda++~\cite{yang2025wandapp} (ACL 2025) extends Wanda with improved pruning metrics.
Our contribution is orthogonal to these methods: we provide a principled \emph{explanation} of which components are sensitive and why (early-layer MLP, low covariance rank, heavy-tailed activations), grounded in Lyapunov-style contraction analysis and machine-checked per-matrix bounds.
The structural heuristics derived from this analysis are competitive with OWL on steep-hierarchy architectures (Qwen3.5-2B) but do not improve over ATP on flat-hierarchy architectures (LLaMA-2-7B) at matched sparsity (Table~\ref{tab:optimal}).
The key contribution is the explanatory framework---predicting \emph{which} architectures benefit from non-uniform allocation (steep per-component sensitivity spread) and \emph{why} (convexity of the per-row error function at high sparsity)---rather than a universal compression method.

\paragraph{Adaptive inference}
Early exit~\cite{schwartz2020right}, token pruning~\cite{kim2022learned}, and speculative decoding~\cite{leviathan2023fast} reduce compute adaptively.
Our approach operates at the weight matrix level rather than the token or layer level, and provides per-matrix error tracking with formal bounds.

\section{Per-Matrix Error Bounds}
\label{sec:framework}

Each weight matrix in a transformer layer performs a linear map $y = Wx$, and we model these as bounded linear maps $f: E \to F$ between normed spaces, proving error bounds for the compression stages applied to each one. These guarantees cover the linear components only---the matrix multiplications themselves---so error composition across the nonlinear components is instead measured empirically via contraction analysis (\S\ref{sec:lyapunov}).

\paragraph{Core bound: sparse approximation (Theorem~2 in \ref{app:lean})}
We decompose $f = f_{\text{sparse}} + f_{\text{residual}}$ where $f_{\text{sparse}}$ retains only the top-$k$\% of weights by magnitude.
The sparse activation bound states:
\[
\|f(x) - f_{\text{sparse}}(x)\| \leq \|f_{\text{residual}}\| \cdot \|x\|
\]
This is the bound used throughout the paper: each pruned weight matrix has a verified error ceiling given by the operator norm of the removed weights.
The error decreases monotonically with the retention fraction $k$.

\paragraph{Additional stages (\ref{app:lean})}
For completeness, we formalize two additional composable stages---semantic caching (Thm.~1 in \ref{app:lean}: $\|f(x) - f(x_c)\| \leq \|f\| \cdot \delta$) and low-rank approximation (Thm.~3: $\|f_{\text{sparse}}(x) - f_{\text{approx}}(x)\| \leq \sigma_{r+1} \cdot \|x\|$).
The three-stage composition bound (Thm.~4) shows the total error is additive.
These are not empirically evaluated in this paper; the experimental focus is pruning (Thm.~2) and layer removal.

\section{Machine-Checked Bounds in Lean 4}
\label{sec:lean}

All theorems in \S\ref{sec:framework} are machine-checked in Lean~4~\cite{moura2021lean} using the Mathlib library's normed space infrastructure.
Their value is \emph{verified local error accounting}: ensuring that the per-matrix bounds used throughout the paper are exactly correct.

\subsection{Proof Architecture}

We represent neural network layers as bounded linear maps \texttt{E →L[ℝ] F} in Mathlib's \texttt{Analysis.NormedSpace} module.
This provides the operator norm $\|f\|$ with the key inequality $\|f(x)\| \leq \|f\| \cdot \|x\|$, linearity $f(x - y) = f(x) - f(y)$, and standard norm properties (non-negativity, triangle inequality, scaling).

Each proof follows a common pattern: rewrite the error using linearity, apply the operator norm bound, then chain inequalities using \texttt{calc} blocks.

\subsection{Proof Strategy and Multi-Layer Bounds}

The composition proof (Theorem~4) rewrites the error using both decompositions, applies the triangle inequality, then bounds each term via \texttt{le\_opNorm}---15 lines of Lean tactic code (\ref{app:lean}).
Theorems 7--9 extend to multi-layer propagation: Theorem~7 gives a linear bound ($n \cdot \max_i \epsilon_i$), while Theorem~8 provides a tighter geometric bound when errors contract by $\rho < 1$ per layer.
Theorem~9 formalizes the contraction condition itself (\S\ref{sec:lyapunov}).
Theorem~10 provides a \emph{sufficient condition} for contraction in Pre-LN transformers: if the sublayer gain $g$ and residual alignment $\cos\theta$ satisfy $1 + 2g\cos\theta + g^2 > (1+L)^2$, where $L$ bounds the error amplification, then $\rho < 1$.
Theorem~11 provides a composition template for layerwise perturbation propagation: if the $i$-th layer has perturbation-growth coefficient $K_i$ and its matrix approximation introduces error $\epsilon_i$, then the total output error is bounded by:
\begin{equation}
\|\text{output}_{\text{approx}} - \text{output}_{\text{exact}}\| \leq \sum_{i=1}^{L} \epsilon_i \prod_{j>i} K_j
\label{eq:composition}
\end{equation}
This provides a composition route from local per-matrix errors to a diagnostic end-to-end estimate: per-matrix errors (Thm.~2) compose through measured layer-level norm-growth ratios.
The estimate is conservative ($\prod K_j$ grows with depth), motivating the empirical $\rho$ measurement as a tighter practical alternative.
\emph{Important caveat:} $K_i = \|h_{t+1}\|/\|h_t\|$ is a trajectory-level norm-growth ratio, not a true Lipschitz constant (which would bound output change under input perturbation via the Jacobian operator norm).
The resulting product should be interpreted as an empirical structural characteristic, not a worst-case guarantee.
Theorem~11 formalizes the 2-layer base case; the $L$-layer bound (Eq.~\ref{eq:composition}) follows by induction.
All twelve theorems type-check without \texttt{sorry}.

\paragraph{Scope and architecture-agnosticism}
Our theorems bound errors in \emph{individual matrix multiplications}, which are exactly linear operations.
The per-matrix bound $\|Wx - \hat{W}x\| \leq \|W_{\text{residual}}\|_2 \cdot \|x\|$ is tight, assumption-free, and \emph{architecture-agnostic}: it makes no assumptions about activation functions, normalization, positional encoding, or attention mechanisms.
This is a deliberate design choice.
The bounds hold identically on GPT-2 Small (2019: GELU, learned positional embeddings) and Qwen3.5-2B (2025: SwiGLU, rotary positional embeddings, grouped-query attention, RMSNorm)---as expected, since the bounds apply to matrix multiplications regardless of surrounding nonlinearities.
Any future transformer variant that uses matrix multiplications inherits the same guarantee without modification.

\paragraph{Bound tightness in practice}
When the per-matrix bounds are composed through Theorem~11, the cumulative norm-growth product $\prod K_i$ is non-vacuous ($< 530$ for models up to 7B, CV $< 4\%$ across inputs), confirming that residual connections fundamentally tighten error composition. Instantiating Theorem~11 with measured ratios gives total error $\lesssim 1\text{--}17$---an empirical diagnostic, not a worst-case guarantee. Theorem~12 connects contraction ($\rho$) to the composition estimate: $\rho < 1$ always tightens the bound. Full norm-growth measurements, bridge analysis, and bound tightness data appear in \ref{app:normgrowth}.

\section{Experimental Setup}
\label{sec:setup}

\paragraph{Model}
GPT-2 Small~\cite{radford2019language}: 12 transformer layers, 768-dimensional hidden state, 12 attention heads (64-dim each), 3072-dimensional MLP intermediate.
Each layer contains 39 weight matrices after splitting \texttt{c\_attn} into per-head Q/K/V projections, yielding 468 total matrices.
We group these into 72 ``compression groups'': 6 component types $\times$ 12 layers.

\paragraph{Data}
WikiText-103 test set~\cite{merity2017pointer} as our primary benchmark.
GPT-2 sensitivity experiments use 2{,}000 tokens (2 chunks of 1{,}024), sufficient for stable per-group perplexity comparisons.
Cross-model comparisons (Table~\ref{tab:crossmodel}) use 4{,}096 tokens (4 chunks of 1{,}024) via a unified evaluation pipeline for consistency across models with different context lengths.
For calibration-aware experiments (\S\ref{sec:compression}), we use a single 1{,}023-token chunk (baseline perplexity 15.7) to match the evaluation scale of the calibration data.
To validate that our findings are not artifacts of the evaluation scale or dataset, we also evaluate at up to 51{,}200 tokens on WikiText-103 and on C4 validation data~\cite{raffel2020exploring} (\ref{app:scale}).

\paragraph{Implementation}
Full GPT-2 forward pass reimplemented in NumPy with \texttt{ApproxLinear} wrappers around each weight matrix.
Each wrapper tracks both the actual approximation error $\|Wx - \hat{W}x\|$ and the theoretical bound $\|W_{\text{residual}}\| \cdot \|x\|$ at every forward pass, enabling violation detection.
Conv1D weights are transposed from HuggingFace format (in, out) to standard (out, in).
SVDs are precomputed and cached per sparsity level.

\paragraph{Configurations}
We evaluate three experimental regimes.
The \emph{ablation study} (67 configs) covers single-layer (12), component-type (6), cumulative forward/backward (24), sparsity-only (5), rank-only (6), and sensitivity sweep on layer~0 (13), with default compression at sparsity=5\%, rank=16.
The \emph{greedy bandit} (36 rounds + baseline) performs sequential group compression at sparsity=5\%, rank=32.
The \emph{verification} study (72 configs) tests cliff robustness across 5 compression levels.

\section{Results}
\label{sec:results}

\begin{table*}[t]
\centering
\caption{Cross-model validation summary. All 50\% Wanda results use a unified evaluation pipeline (weight$\times$column-norm scoring, 4{,}096 tokens, WikiText-103 test). Lyapunov measurements use $\epsilon=0.01$ perturbation on 512 tokens. ``Contracting'' reports the number of inter-layer transitions (out of $L{-}1$) with $\rho < 1$. ``---'' indicates per-matrix bounds were not measured for that model.}
\label{tab:crossmodel}
\footnotesize
\begin{tabular}{lrrrrrr}
\toprule
& GPT-2 S & GPT-2 M & Qwen3.5 & LFM2 & Mistral & Qwen3 \\
\midrule
Parameters & 117M & 345M & 1.9B & 2.6B & 7.2B & 8B \\
Layers & 12 & 24 & 24 & 30 & 32 & 36 \\
$d_{\text{model}}$ & 768 & 1024 & 2048 & 2560 & 4096 & 4096 \\
Architecture & Transf. & Transf. & Transf. & Hybrid & Transf. & Transf. \\
Baseline PPL & 26.53 & 19.90 & 13.54 & 34.15 & 6.31 & 10.38 \\
\midrule
\multicolumn{7}{l}{\textit{Lyapunov stability}} \\
Contracting & 11/11 & 21/23 & 13/23 & 19/29 & 17/31 & 18/35 \\
$\rho_{\max}$ & 0.96 & 2.05 & 1.11 & 1.12 & 1.14 & 1.33 \\
\midrule
\multicolumn{7}{l}{\textit{50\% Wanda pruning}} \\
Degradation & $120\times$ & $122\times$ & $3\times$ & $7\times$ & $11\times$ & $42{,}922\times$ \\
$\tau$ & 0.50 & 0.50 & 0.60 & 0.80 & 0.95 & 0.80 \\
Violations & 0 & 0 & 0 & --- & 0 & --- \\
\bottomrule
\end{tabular}
\end{table*}

Table~\ref{tab:crossmodel} summarises results across all six architectures. Across three models---GPT-2 Small (14{,}040 per-matrix checks), Qwen3.5-2B (1{,}744{,}896), and Mistral-7B (2{,}899{,}968)---we observe \textbf{zero bound violations}, confirming correct implementation of the per-matrix bounds in IEEE 754 arithmetic. Zero violations are \emph{expected} given the bound construction (operator norm bounds are tight by definition); the check serves as an implementation sanity test, not an empirical discovery. The result holds at 51{,}200 tokens on WikiText-103 and 10{,}000 tokens on C4; detailed per-layer and per-component data appear in \ref{app:ablation} and \ref{app:scale}.

\subsection{Structural Sensitivity Hierarchy}
\label{sec:sensitivity}

We use two metrics: \emph{regret} (additive PPL increase) for per-component comparisons, and \emph{degradation factor} (PPL ratio) for cross-model comparisons where baselines differ.

Uniform compression fails: the best uniform configuration reaches perplexity 3{,}865 ($168\times$ baseline),\footnote{GPT-2 sensitivity experiments use the NumPy reimplementation (baseline 23.0). Table~\ref{tab:crossmodel} reports 26.53 from the unified HF pipeline on 4{,}096 tokens. The difference reflects tokenization and chunk count; each methodology is internally consistent.} forcing non-uniform allocation.

\begin{figure*}[t]
\centering
\includegraphics[width=\textwidth]{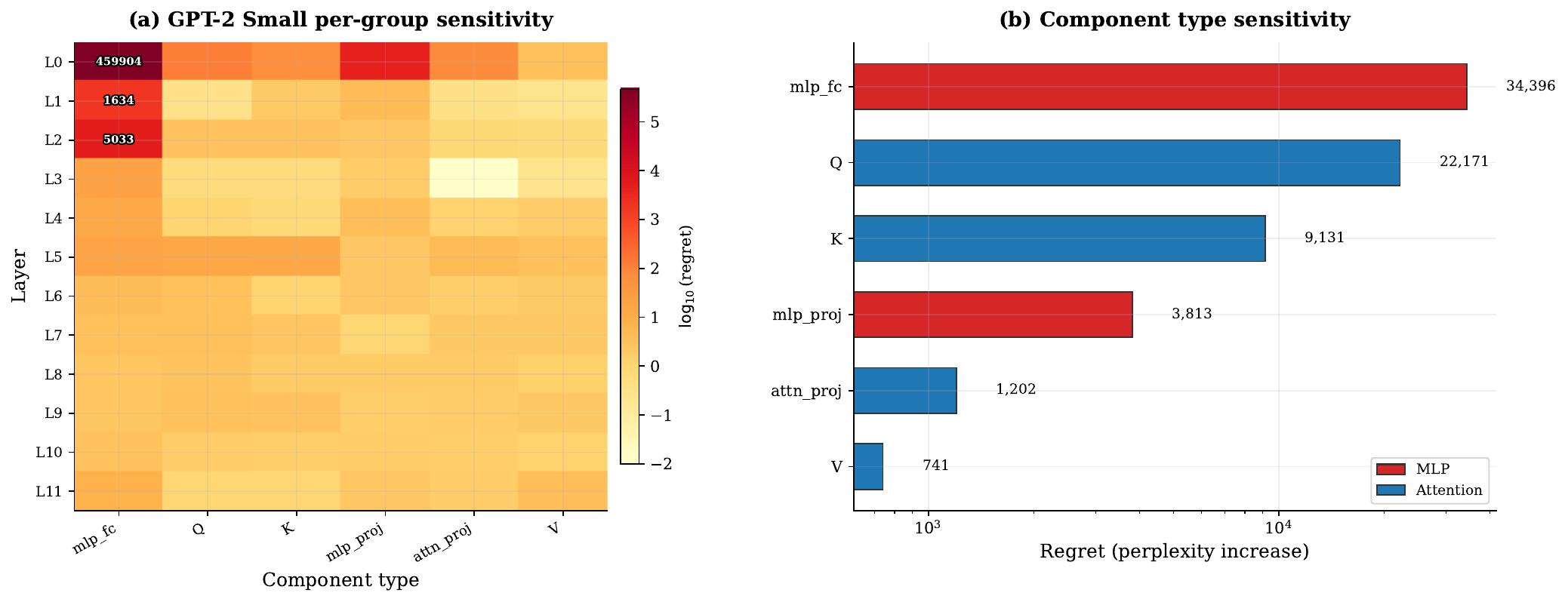}
\caption{Structural sensitivity analysis. (a)~Per-group compression regret on log scale ($12 \times 6$ heatmap). Each cell shows the perplexity increase when that group alone is compressed (sparsity=5\%, rank=32). Layer~0 \texttt{mlp\_fc} dominates at 459{,}904. (b)~Component-type sensitivity hierarchy: MLP components (red) are consistently more sensitive than attention (blue), spanning a $46\times$ range.}
\label{fig:heatmap}
\end{figure*}

\paragraph{Five orders of magnitude}
Single-layer ablation (compressing one layer at sparsity=5\%, rank=16) reveals that layer~0 is five orders of magnitude more sensitive than mid-layers: perplexity 1{,}539{,}140 vs.\ 26--29 for layers 6--10 (supplementary material).
Component-type ablation shows a consistent hierarchy: \texttt{mlp\_fc} $\gg$ Q $>$ K $>$ \texttt{mlp\_proj} $>$ \texttt{attn\_proj} $>$ V, spanning a 46$\times$ range from regret 34{,}396 to 741 (supplementary material).
Both hierarchies are confirmed at 51{,}200 tokens on WikiText-103 and at 10{,}000 tokens on C4 (supplementary material).
Figure~\ref{fig:heatmap} visualizes the full 72-group sensitivity map.

\begin{figure*}[t]
\centering
\includegraphics[width=\textwidth]{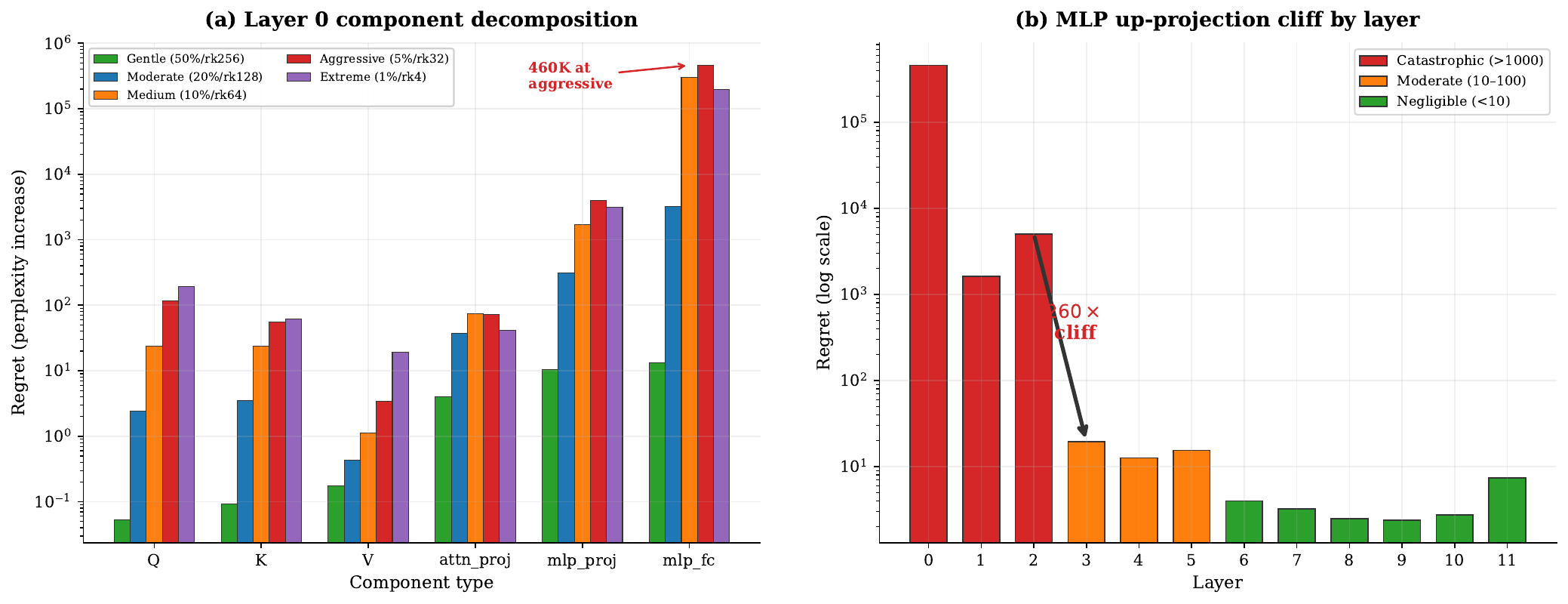}
\caption{The early-layer MLP catastrophe. (a)~Layer~0 component decomposition across five compression levels: \texttt{mlp\_fc} dominates at every level, reaching 460K regret at aggressive compression. V projections remain nearly free even at extreme settings. (b)~\texttt{mlp\_fc} regret by layer reveals a sharp phase transition between layers 2 and 3---a ${\sim}250\times$ cliff separating catastrophic from negligible sensitivity.}
\label{fig:mlp_cliff}
\end{figure*}

\paragraph{The early-layer MLP catastrophe}
Compressing layer~0's \texttt{mlp\_fc} alone---one matrix out of 468---increases perplexity from 23.0 to 459{,}927 ($20{,}000\times$).
This sensitivity drops sharply between layers 2 and 3: a ${\sim}250\times$ cliff separates catastrophic (regret $>1{,}000$) from negligible (regret $<4$) layers (Figure~\ref{fig:mlp_cliff}b).
The cliff persists across all five compression levels tested (Figure~\ref{fig:mlp_cliff}a; data in supplementary material), and generalizes to GPT-2 Medium where layer~0 \texttt{mlp\_fc} yields PPL 315{,}029 vs.\ layer~1's 20.9---a $15{,}000\times$ cliff.

The cliff position is invariant across all five compression levels tested, and is associated with three correlated properties of early layers (low output covariance rank, heavy-tailed activations, low GELU sparsity); per-layer measurements are reported in the supplementary material. The absolute degradation magnitudes are inflated by our simplified Wanda (weight$\times$column-norm scoring without calibration data); activation-aware methods~\cite{sun2024wanda} produce smaller absolute numbers, but the structural pattern is the contribution, not the absolute scale.

\begin{figure}[t]
\centering
\includegraphics[width=0.85\linewidth]{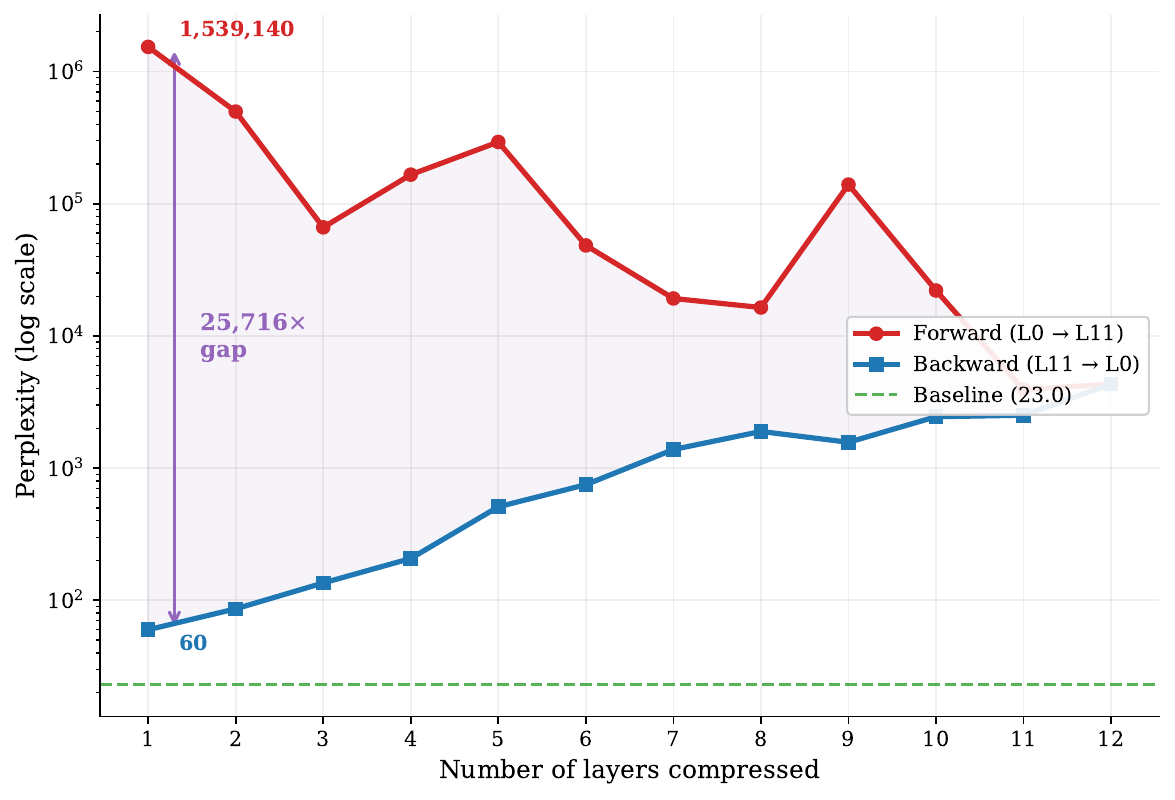}
\caption{Cumulative compression: forward (L0$\to$L11) vs.\ backward (L11$\to$L0). The $25{,}716\times$ gap at one layer compressed demonstrates that layer~0 dominates overall sensitivity.}
\label{fig:forward_backward}
\end{figure}

\paragraph{Forward/backward asymmetry}
Layer ordering is critical (Figure~\ref{fig:forward_backward}). Compressing from L0 forward yields PPL 1.54M after just one layer, while compressing backward from L11 yields only 60---a $25{,}716\times$ gap, all driven by layer~0's outsized sensitivity. Excluding early layers from compression therefore yields large efficiency gains.

\subsection{Lyapunov-Style Contraction Analysis}
\label{sec:lyapunov}

The structural sensitivity results raise a fundamental question: \emph{why} do compression errors not accumulate catastrophically through 12 layers?
We address this by measuring relative error energy inspired by Lyapunov stability theory~\cite{khalil2002nonlinear,haber2017stable}.
We emphasize that our contraction factors $\rho_t$ are \emph{empirically measured} along specific input trajectories, not global stability certificates---they characterize typical behavior, not worst-case guarantees.

Define the relative error energy at layer $t$:
\begin{equation}
V(t) = \frac{\|e_t\|^2}{\|h_t\|^2}, \quad e_t = h_t - \hat{h}_t
\end{equation}
where $h_t$ is the true hidden state and $\hat{h}_t$ the approximate state.
If each layer contracts errors, $V(t+1) \leq \rho_t \cdot V(t) + c_t$ with contraction factor $\rho_t < 1$ and per-layer injection $c_t$ from the compression error.
In practice, we measure $\rho_t = V(t{+}1)/V(t)$ directly, which absorbs $c_t$; the measured ratio captures total error evolution without separating contraction from injection.

\paragraph{LayerNorm geometry}
The Jacobian of LayerNorm at point $z$ with mean $\mu$ and variance $\sigma^2$ is:
\begin{equation}
J_{\text{LN}} = \frac{1}{\sigma}\left(I - \frac{1}{d}\mathbf{1}\mathbf{1}^\top - \frac{(z-\mu)(z-\mu)^\top}{d\sigma^2}\right)
\end{equation}
This projects out the mean and radial components, with spectral radius $\|J_{\text{LN}}\|_2 \leq 1/\sigma$---perturbations aligned with the current activation direction are contracted.
If each layer's effective gain is less than unity, bounded per-layer perturbations produce bounded total error.

\paragraph{Empirical measurement}
We inject controlled perturbations ($\epsilon \in \{0.001, 0.01, 0.05, 0.1\}$) at the embedding layer and track $V(t)$ through all layers. For GPT-2 Small, all 11 inter-layer transitions contract errors ($\rho_{\max} = 0.96$), with layer~0 providing the strongest contraction ($\rho_0 \approx 0.25$ at $\epsilon = 0.001$). The contraction pattern is stable across perturbation magnitudes, and the resulting geometric-series bound is $23\times$ tighter than the linear $n \cdot \max_i \epsilon_i$ alternative. GPT-2 Medium contracts in 21 of 23 inter-layer transitions, though the final layer amplifies ($\rho_{23} \approx 2.05$); the overall trajectory still shows net contraction through the interior.

\paragraph{Why contraction occurs: the residual growth mechanism}
The contraction of $V(t)$ is \emph{not} primarily due to LayerNorm damping the error---in fact, the operator norm of each sub-layer's Jacobian far exceeds 1 (we measure $L_f/\sigma \approx 300\text{--}13{,}000$).
Instead, contraction arises from the interplay between error growth and \emph{hidden state growth} through residual connections.

In a Pre-LN (LayerNorm-before-sublayer) transformer, $h_{t+1} = h_t + f(\text{LN}(h_t))$.
The relative error evolves as:
\begin{equation}
\label{eq:contraction}
\rho_t = \frac{V(t{+}1)}{V(t)} = \underbrace{\left(\frac{\|e_{t+1}\|}{\|e_t\|}\right)^2}_{\text{error amplification } \beta^2} \cdot \underbrace{\left(\frac{\|h_t\|}{\|h_{t+1}\|}\right)^2}_{\text{denominator growth } 1/\alpha^2}
\end{equation}
Contraction ($\rho_t < 1$) occurs when the hidden state grows faster than the error: $\alpha > \beta$, yielding $\rho_t \leq (\beta/\alpha)^2$.

Figure~\ref{fig:deep_lyapunov} validates this decomposition on GPT-2 Small and Mistral-7B (per-layer data in supplementary material).
Layer~0 contracts most ($\rho = 0.25$) because the embedding-to-hidden expansion is $\sim 11\times$ ($\alpha^2 = 123$), massively diluting perturbations.
Late layers contract strongly ($\rho_{10{-}11} \approx 0.65$) as residual accumulation grows $\|h_t\|$.
GPT-2 Medium layer~23 amplifies ($\rho \approx 2.0$) because its attention sub-layer produces a disproportionately large output, growing error faster than the hidden state grows.

\begin{figure*}[t]
\centering
\includegraphics[width=\textwidth]{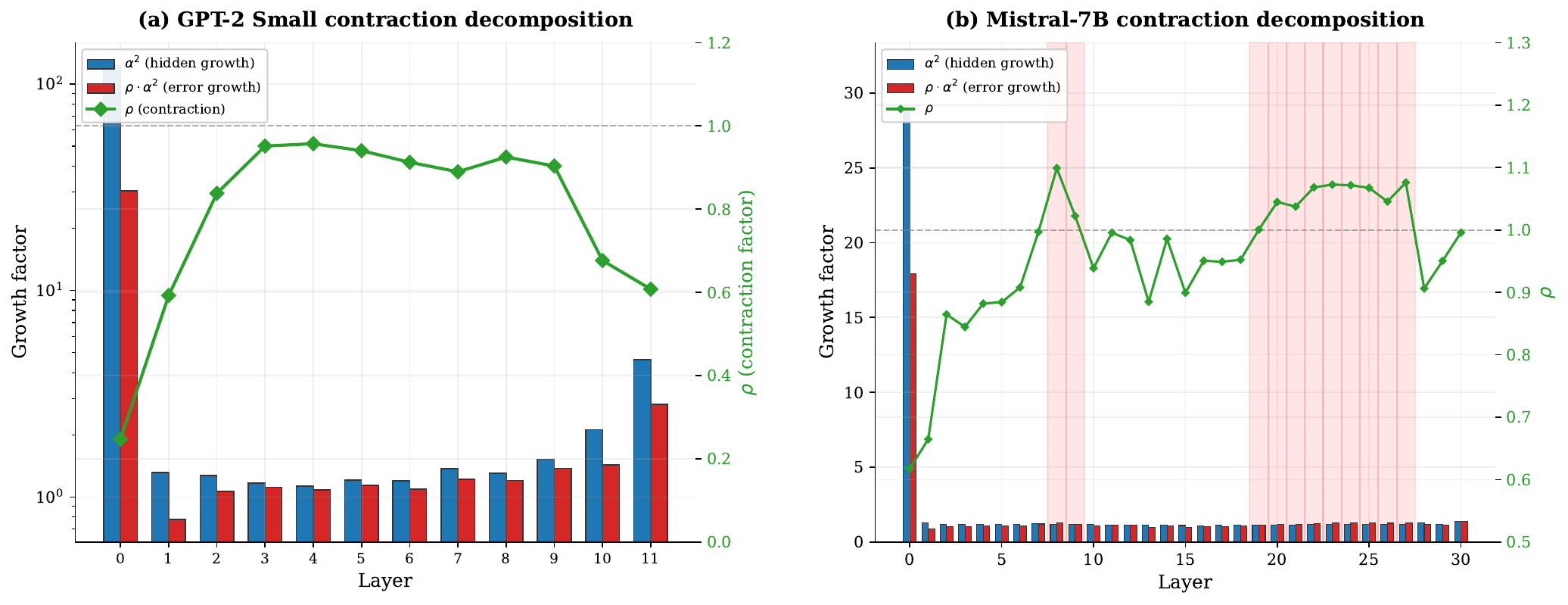}
\caption{Contraction decomposition. (a)~GPT-2 Small: blue bars show hidden state growth ($\alpha^2$), red bars show error growth ($\rho \cdot \alpha^2$), green diamonds show contraction factor $\rho$. Layer~0's massive expansion ($\alpha^2 = 123$) produces the strongest contraction ($\rho = 0.25$). All layers contract. (b)~Mistral-7B across 31 layers. Red-shaded regions highlight amplifying layers ($\rho > 1$).}
\label{fig:deep_lyapunov}
\end{figure*}

\paragraph{Formal verification}
Two Lean~4 theorems formalize the Lyapunov analysis.
Theorem~8 establishes the geometric series bound: if $\rho < 1$, then $c_{\max}(1 - \rho^n)/(1 - \rho) \leq c_{\max}/(1 - \rho)$, giving a tighter bound than the linear $n \cdot \max$ for all models where $\rho < 1$.
Theorem~9 formalizes the contraction condition: if the hidden state growth factor $\alpha$ exceeds the error growth factor $\beta$ (i.e., $\beta < \alpha$), then $(\beta/\alpha)^2 < 1$, guaranteeing contraction.
Both are machine-checked with no \texttt{sorry}, formalizing the multi-layer error bound \emph{conditional on} the empirically measured contraction factor $\rho$.

\paragraph{Connection to the early-layer catastrophe}
The Lyapunov analysis also sheds light on why early-layer compression is catastrophic.
Layer~0 provides the strongest error contraction ($\rho_0 \approx 0.25$), meaning it normally \emph{dampens} errors introduced at the embedding level.
When layer~0's MLP is itself corrupted, the error is introduced \emph{after} the contractive mechanism, and all subsequent layers must absorb an error that bypassed the strongest dampening stage.
The low output covariance rank of early MLP layers means the error occupies dimensions that subsequent LayerNorms cannot project away.

\subsection{Cross-Architecture Validation at Scale}
\label{sec:crossarch}

Having explained the contraction mechanism on GPT-2 Small, we now examine how contraction, sensitivity, and compression tolerance vary across six architectures (Table~\ref{tab:crossmodel}).

\paragraph{Per-model findings}
Key findings per model (full details in supplementary):
\textbf{GPT-2 Medium} (345M): early-layer catastrophe generalizes (L0 PPL $= 874{,}632$); \texttt{mlp\_fc} cliff is sharper (15{,}000$\times$ drop from L0 to L1); zero bound violations.
\textbf{Qwen3.5-2B}: MLP $>$ attention ordering preserved with both simplified and activation-aware Wanda; $\tau = 0.60$; zero violations across 1.74M checks.
\textbf{LFM2-2.6B} (hybrid): contraction operates in hybrid architectures ($\rho_{\max} = 1.12$); most compression-resilient model ($7\times$ degradation at 50\%), suggesting conv-attention redundancy buffers perturbations.
\textbf{Mistral-7B}: no catastrophic single-layer sensitivity even at 90\% sparsity (worst regret $+33$, vs.\ GPT-2's $459{,}904$ at 5\%); V-projection ranking is method-dependent (simplified vs.\ activation-aware Wanda).

\paragraph{Catastrophe threshold scaling}
The early-layer catastrophe is not absent at scale---it merely requires more aggressive compression.
We define the catastrophe threshold $\tau$ as the maximum sparsity at which single-layer Wanda pruning of layer~0 yields regret $< 100$.
Sweeping all six models from 50\% to 99\% sparsity on layer~0, we find that $\tau$ generally increases with $d_{\text{model}}$ but is modulated by architecture (Table~\ref{tab:threshold_all}):

\begin{table}[!htbp]
\centering
\caption{Catastrophe threshold $\tau$ across six architectures (layer-0 Wanda, regret $< 100$).}
\label{tab:threshold_all}
\small
\begin{tabular}{lrrrl}
\toprule
Model & $d$ & $\tau$ & $1{-}\tau$ & Note \\
\midrule
GPT-2 Small & 768 & 0.50 & 0.50 & Catastrophic at 70\% \\
GPT-2 Medium & 1024 & 0.50 & 0.50 & Same as Small \\
Qwen3.5-2B & 2048 & 0.60 & 0.40 & Full per-matrix study \\
LFM2-2.6B & 2560 & 0.80 & 0.20 & Hybrid architecture \\
Qwen3-8B & 4096 & 0.80 & 0.20 & Same $\tau$ as LFM2 \\
Mistral-7B & 4096 & 0.95 & 0.05 & Most resilient \\
\bottomrule
\end{tabular}
\end{table}

Dimension alone does not determine $\tau$: Qwen3-8B ($d = 4096$) has $\tau = 0.80$, the same as LFM2-2.6B ($d = 2560$), while Mistral-7B at the same dimension achieves $\tau = 0.95$.
This confirms that architecture-specific factors---Lyapunov stability ($\rho_{\max}$), tied embeddings, and hybrid design---modulate the threshold independently of width, consistent with the Compression Fragility Index (CFI) analysis in \S\ref{sec:discussion}.

\begin{figure}[t]
\centering
\includegraphics[width=0.85\linewidth]{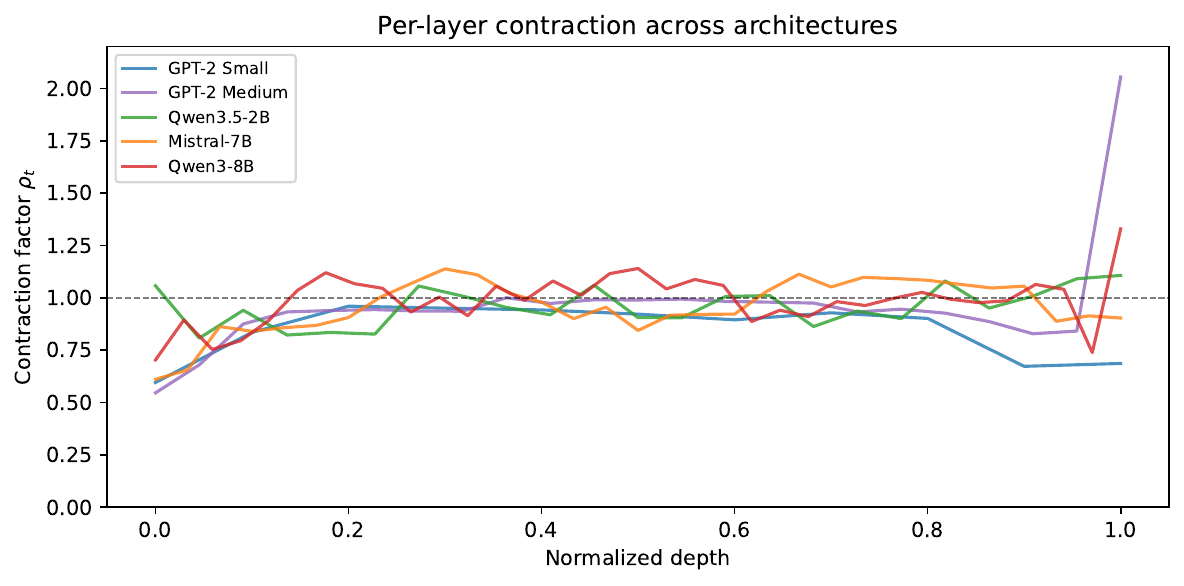}
\caption{Per-layer contraction factor $\rho_t$ across six architectures on a normalized depth axis. GPT-2 Small (blue) contracts everywhere ($\rho < 1$). LFM2 (green), Mistral (orange), and Qwen3 (red) all cross the stability boundary, with GPT-2 Medium (purple) spiking to $\rho = 2.05$ at the final layer.}
\label{fig:rho_comparison}
\end{figure}

\paragraph{Contraction is necessary but not sufficient}
Among similar architectures, higher $\rho_{\max}$ correlates with worse outcomes: Qwen3-8B ($\rho_{\max} = 1.33$) collapses at $42{,}922\times$ while Mistral-7B ($\rho_{\max} = 1.14$) degrades only $11\times$.
However, GPT-2 Small ($\rho_{\max} = 0.96$, fully contracting) degrades $120\times$, while LFM2-2.6B ($\rho_{\max} = 1.12$) degrades only $7\times$ (Table~\ref{tab:crossmodel}).
Compression tolerance depends on both error stability and width-dependent redundancy ($1/\sqrt{d}$ scaling).
Figure~\ref{fig:rho_comparison} shows the contraction profiles; per-model details and the Qwen3-8B analysis appear in \ref{app:crossarch}.

Activation-aware Wanda (with calibration data) confirms that MLP $>$ attention sensitivity ordering is structural, not an artifact of simplified scoring: the ordering is preserved across GPT-2 Small, Qwen3.5-2B, and Mistral-7B (full per-component data in \ref{app:crossarch}).

\paragraph{Comparison with state-of-the-art}
Unlike SparseGPT~\cite{frantar2023sparsegpt} and Wanda~\cite{sun2024wanda}, which focus on pruning quality, our framework explicitly instruments and machine-checks per-matrix residual-error accounting ($\|Wx - W'x\| \leq \|W_{\text{residual}}\|_2 \cdot \|x\|$), then studies how those errors propagate empirically.
The sensitivity hierarchy identified here is orthogonal to the compression method: it characterizes \emph{which} components to protect, a decision that applies equally whether compression is performed via SparseGPT, GPTQ~\cite{frantar2023gptq}, or any other method.
The activation-aware validation (\ref{app:crossarch}) confirms that the MLP $>$ attention ordering is structural, not an artifact of our simplified scoring.

\subsection{Compression Methods and Allocation}
\label{sec:compression}

The structural sensitivity and Lyapunov findings motivate compression methods that respect the hierarchy.

\paragraph{Calibration-aware compression}
We evaluate two approaches: activation-aware pruning (Wanda~\cite{sun2024wanda}) and balanced truncation from control theory~\cite{moore1981principal,glover1984all}.

Wanda achieves $10\times$ lower perplexity than naive sparsification (34.6 vs.\ 406.7 at 50\%), confirming that activation-aware pruning exploits the structural information our analysis identifies.
Protecting layer~0 \texttt{mlp\_fc} provides further improvement (32.3 vs.\ 34.6), consistent with the catastrophe analysis.
Balanced truncation~\cite{moore1981principal,glover1984all} extends the safe frontier by $2.5\times$: at aggressive compression, 7 groups achieve PPL $<30$ vs.\ 3 for naive SVD.

\begin{figure*}[t]
\centering
\includegraphics[width=\textwidth]{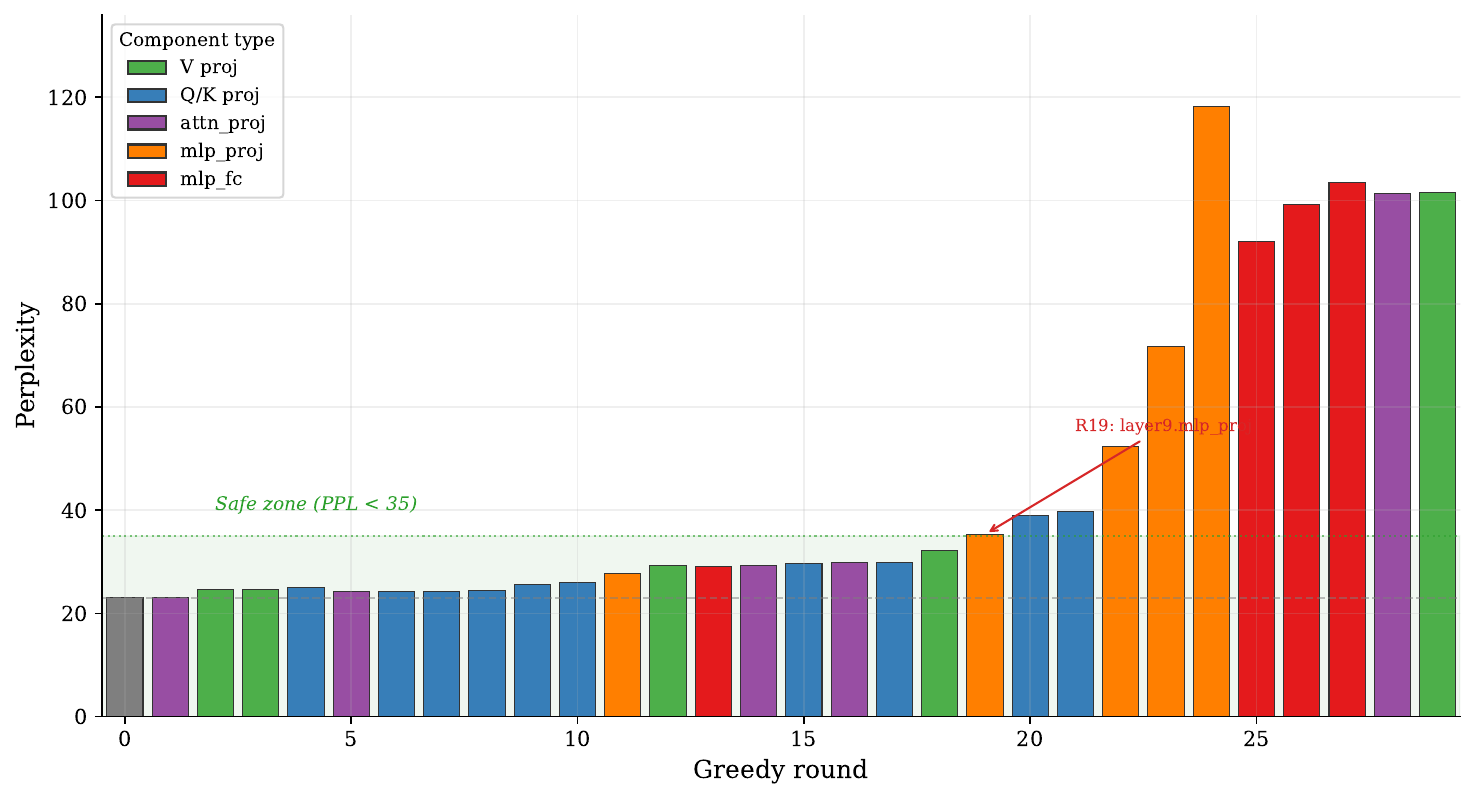}
\caption{Greedy compression waterfall on GPT-2 Small. V/attention outputs (green, purple) compress first at near-zero cost; early-layer MLPs (orange) trigger the exit from the safe zone.}
\label{fig:greedy}
\end{figure*}

\paragraph{Greedy allocation and contraction budget}
A greedy strategy (sequentially compressing the lowest-regret group) achieves 24\% FLOPS savings at +27\% perplexity (Figure~\ref{fig:greedy}).
The greedy ordering naturally rediscovers the sensitivity hierarchy: V projections first, early-layer MLPs last.
The Lyapunov budget ($A_t = \prod_{s>t} \rho_s$) allocates 54\% of error tolerance to layers 8--11 and only 19\% to layers 0--3, but correlates weakly with greedy ordering (Spearman $r = 0.25$) because within-layer component sensitivity dominates.

However, the Lyapunov \emph{budget}---defined as the cumulative downstream contraction $1/\prod_{s>t} \rho_s$, measuring how much error each layer position can tolerate---correlates strongly with layer-averaged compression regret (Spearman $r = 0.85$, $p < 0.001$).
This correlation is partially confounded by the early-layer catastrophe (both budget and regret are highest for early layers), but confirms that the contraction profile captures the correct layer-level ordering, while component-level sensitivity requires separate empirical measurement.
Effective compression allocation therefore requires two complementary tools: Lyapunov budgets for \emph{inter-layer} error tolerance, and component-level sensitivity scores for \emph{intra-layer} allocation.

\paragraph{Analytic contraction-based schedule}
The inter-layer allocation has a closed-form solution under a quadratic error model: $\beta^* = -\bar{s} \cdot \sum_t w_t d_t / \sum_t w_t d_t^2$ where $w_t = \|W_t\|_F^2 \cdot A_t$ (weight norms $\times$ Lyapunov amplification, zero search cost).
The formula correctly predicts the schedule \emph{direction} but overestimates the \emph{magnitude} ($\beta^* = 0.031$ vs.\ ATP's searched $0.004$ at 50\%) because the actual error function is subquadratic.
When clamped, the Lyapunov schedule matches ATP: PPL 20.2 vs.\ 20.1 at 70\% on LLaMA-2-7B.
Full derivation and analysis are in supplementary material.

\paragraph{Practical compression guidelines}
Distilling the sensitivity analysis across six models and two pruning methods yields three architecture-agnostic rules. First, protect early-layer MLP up-projections: never compress the first $\lfloor L/8 \rfloor$ layers' MLP gate/up-projections, which operate at near-full rank (regret $+27$ to $+460{,}000$ depending on scale).
Second, compress K/V projections first: these consistently have the lowest regret ($< 0.3$ at 50\% sparsity with activation-aware Wanda).
Third, use the catastrophe threshold $\tau$ to set the sparsity ceiling ($\tau$ ranges from 0.50 for GPT-2 Small to 0.95 for Mistral-7B); a single contraction measurement plus one layer-0 sweep suffices to estimate $\tau$.
These rules require no model-specific tuning.

\paragraph{Allocation results (Table~\ref{tab:optimal})}
Rule-based allocation wins on 1 of 3 architectures (Qwen3.5-2B: PPL 23.7 vs.\ Wanda 26.0 and OWL 31.6) but trails Wanda on Mistral-7B (8.8 vs.\ 8.4) and clusters with all methods on LLaMA-2-7B at 50\%.
The critical factor is the sensitivity spread: architectures with steep hierarchies benefit from non-uniform allocation, while flat hierarchies ($3\text{--}7\times$ under activation-aware Wanda) are better served by uniform methods like ATP~\cite{zhuang2025atp}.

More striking is the within-layer story: reallocation fails on every architecture tested---including Mamba-370M ($10^{13}\times$ simplified-pruning spread).
At matched per-layer sparsity, every redistribution strength $\alpha > 0$ hurts (PPL 6.88 $\to$ 10.56 at 50\%).
The root cause is convexity of $E(s)$: per-row Wanda already acts as a strong implicit within-layer allocator.
ATP's success comes from its \emph{inter-layer} schedule, which the contraction analysis explains: $A_t = \prod_{s>t} \rho_s$ decreases with depth, making early-layer protection essential.

V-projection sensitivity reverses between GPT-2 (least sensitive) and LLaMA-2 (second most sensitive), demonstrating that generic rules derived from one architecture can harm another.
Full comparison data, analytic schedule derivation, and per-architecture details appear in \ref{app:compression}.

\subsection{Layer Removal}
\label{sec:layerremoval}

Since within-layer reallocation fails, we test a different lever: \emph{entire layer removal}, guided by the empirical contraction factor $\rho$.
Layers with $\rho \approx 1$ neither amplify nor dampen relative perturbation energy.
While $\rho \approx 1$ does not strictly imply the layer is near-identity (a layer could transform representations while preserving relative error), we find that $|\rho - 1|$ correlates with removal regret (Spearman $r = 0.46$, $p = 0.01$), confirming its utility as a removability proxy.
On LLaMA-2-7B (32 layers), single-layer removal ablation confirms this: 21 of 31 testable layers can be individually removed with $<0.5$ PPL regret, with L12 ($\rho = 1.043$, closest to 1.0) being the most removable ($+0.22$ PPL).
The catastrophic layers (L0 at $\rho = 9.41$, L1 at $\rho = 1.13$) are correctly identified by their distance from 1.0; Spearman correlation between $|\rho - 1|$ and removal regret is $r = 0.46$ ($p = 0.01$).

Progressive removal (Table~\ref{tab:layer_removal}, hook-based skipping) shows that $|\rho - 1|$ ranking consistently achieves lower perplexity than both the from-end heuristic and ShortGPT's Block Influence (BI) metric~\cite{men2024shortgpt}: at 4 layers removed, $|\rho{-}1|$ achieves PPL 7.9 vs.\ BI 8.5 and from-end 10.0; at 8 layers, 14.6 vs.\ BI 23.8 ($1.6\times$) and from-end 25.4 ($1.7\times$).
Our BI implementation reproduces ShortGPT's published layer ranking exactly (8/8 overlap with their reported order $\{27,26,25,28,24,29,23,21\}$).
The two metrics select completely non-overlapping layers (0/8 overlap): BI removes deep layers (L21--L29), while $|\rho - 1|$ selects scattered middle layers (L7--L14) that are empirically low-regret.
The advantage of $|\rho - 1|$ grows with removal rate because deep-layer removal (BI) progressively destroys the output pathway, while middle-layer removal (Lyapunov) preserves both the embedding and output stages.

\textbf{Weight pruning gives no inference speedup.}
A critical practical finding: ATP 50\% weight pruning achieves $0.96\times$ speed (Table~\ref{tab:downstream})---\emph{slower} than dense inference on commodity GPUs without sparse tensor cores.
Only physical layer removal translates parameter reduction to wall-clock speedup ($1.22\times$ at 4 layers).
Combining layer removal with ATP is counterproductive ($0.78\times$): the overhead of processing sparse weights in the remaining layers negates the speedup from fewer layers.
This suggests that for deployment on commodity hardware without sparse tensor cores, layer removal is the only pruning-based strategy among those we test that delivers real speedup.

\textbf{Downstream task evaluation} (Table~\ref{tab:downstream}) reveals a nuance: on zero-shot benchmarks (HellaSwag, ARC-Easy, WinoGrande, PIQA), BI-4 (ShortGPT, 68.2\% avg) slightly outperforms $|\rho{-}1|$-4 (65.1\%), despite $|\rho{-}1|$-4 having better perplexity (7.92 vs.\ 8.45, Table~\ref{tab:layer_removal}).
This suggests deep-layer removal (BI) better preserves task-specific features, while middle-layer removal ($|\rho{-}1|$) better preserves language modeling coherence.

\textbf{Combined criterion.}
We test a blended score $\lambda \cdot \widehat{|\rho{-}1|} + (1{-}\lambda) \cdot \widehat{\text{BI}}$ (both normalized to $[0,1]$) across $\lambda \in \{0, 0.25, 0.5, 0.75, 1.0\}$ with physical layer deletion.\footnote{Physical deletion PPL differs slightly from hook-based removal (Table~\ref{tab:layer_removal}) due to exact KV cache handling; all combined-criterion numbers are from a single consistent run.}
At 4 layers removed, $|\rho{-}1|$ dominates the blend for all $\lambda \geq 0.5$, selecting the same layers.
At 8 layers removed, the blend ($\lambda = 0.5$) \textbf{dominates both pure strategies}: selecting a mix of middle and deep layers [10--14, 24, 26, 30], it achieves PPL 14.2 (vs.\ $|\rho{-}1|$-only 16.2, BI-only 25.3) \emph{and} downstream accuracy 60.0\% (vs.\ $|\rho{-}1|$-only 58.3\%, BI-only 59.9\%)---all within the same experimental run.
The improvement comes from preserving both the output pathway (avoiding BI's deep-layer destruction) and near-identity middle layers (leveraging $|\rho{-}1|$'s contraction insight).
Table~\ref{tab:combined} reports the full LLaMA-2 results; Table~\ref{tab:layer_removal} uses hook-based skipping (slightly different absolute PPL, same ordering).

\textbf{Cross-model validation} (Table~\ref{tab:crossmodel_lr}).
We validate on Mistral-7B and Qwen3.5-2B with the same five selection criteria, plus speed benchmarks on Mistral.
On Mistral-7B at $k{=}4$, BI wins on both PPL (8.2 vs.\ $|\rho{-}1|$'s 9.0) and accuracy (72.1\% vs.\ 70.5\%), because Mistral's most removable layers are deep (L23--L27).
On Qwen3.5-2B, $|\rho{-}1|$ wins on PPL (33.4 vs.\ BI's 46.2).
The blend is never the worst method on any architecture, making it the safest cross-architecture default.
From-end removal is catastrophic on both models (PPL 62 on Mistral, 2{,}467 on Qwen); random removal has massive variance (PPL 8.7--134.7 across 3 seeds on Mistral), confirming principled selection is essential.

Mistral speed benchmarks show $|\rho{-}1|$-guided removal gives $1.06\times$ at $k{=}4$ and $1.25\times$ at $k{=}8$ (vs.\ dense 20.1 tok/s).
BI $k{=}4$ is \emph{slower} than dense ($0.73\times$) because removing deep output-pathway layers disrupts KV cache efficiency.

\textbf{Which metric best predicts removability?}
On LLaMA-2-7B, $|\rho{-}1|$ correlates with single-layer removal regret ($r = 0.46$, $p = 0.01$).
On Mistral-7B, $|\rho{-}1|$ does \emph{not} significantly predict regret ($r = 0.12$, $p = 0.50$), but the residual change $\|h_{t+1} - h_t\|/\|h_t\|$ does ($r = 0.48$, $p = 0.006$).
BI is not significant on either model individually.
This confirms that contraction-based metrics are useful proxies but architecture-dependent; the blend hedges across architectures.

\begin{table}[!htbp]
\centering
\caption{Cross-model layer removal at $k{=}4$ (12.5\% of layers). Best PPL in bold. The winner is architecture-dependent; the blend is competitive everywhere. Random removal has massive variance (Mistral: PPL 8.7--134.7 across 3 seeds).}
\label{tab:crossmodel_lr}
\small
\begin{tabular}{llrrr}
\toprule
Model & Method & PPL & Avg acc. & Speed \\
\midrule
\multirow{5}{*}{LLaMA-2-7B} & $|\rho{-}1|$ & \textbf{8.6} & 65.1\% & $1.22\times$ \\
& BI & 8.8 & \textbf{68.2\%} & $1.18\times$ \\
& Blend & \textbf{8.6} & 65.1\% & --- \\
& From-end & 9.7 & --- & --- \\
& Random & --- & --- & --- \\
\midrule
\multirow{5}{*}{Mistral-7B} & $|\rho{-}1|$ & 9.0 & 70.5\% & $1.06\times$ \\
& BI & \textbf{8.2} & \textbf{72.1\%} & $0.73\times$ \\
& Blend & 8.4 & 71.8\% & $1.06\times$ \\
& From-end & 62.3 & 65.8\% & --- \\
& Random & 52.5$\pm$58 & 40.7\% & --- \\
\midrule
\multirow{3}{*}{Qwen3.5-2B} & $|\rho{-}1|$ & \textbf{33.4} & ---$^\dagger$ & --- \\
& BI & 46.2 & --- & --- \\
& From-end & 2{,}467 & --- & --- \\
\bottomrule
\end{tabular}

\vspace{1mm}
{\footnotesize $^\dagger$Qwen3.5-2B downstream not available (lm\_eval incompatibility). Speed: LLaMA-2 from Table~\ref{tab:downstream}; Mistral measured in this experiment.}
\end{table}

\begin{table}[!htbp]
\centering
\caption{Combined layer-removal criterion on LLaMA-2-7B (physical deletion, single consistent run). At 8 layers, the $\lambda{=}0.5$ blend dominates both pure strategies on PPL \emph{and} downstream accuracy.}
\label{tab:combined}
\small
\begin{tabular}{llrr}
\toprule
$\lambda$ & Layers removed & PPL & Avg acc. \\
\midrule
\multicolumn{4}{l}{\textit{4 layers removed (12.5\%)}} \\
0 (BI only) & [24, 26, 27, 28] & 8.77 & 68.2\% \\
0.25 & [24, 26, 27, 30] & 9.74 & 67.8\% \\
0.5--1.0 ($|\rho{-}1|$) & [10, 12, 13, 30] & 8.63 & 65.1\% \\
\midrule
\multicolumn{4}{l}{\textit{8 layers removed (25\%)}} \\
0 (BI only) & [21, 23--29] & 25.3 & 59.9\% \\
\textbf{0.5 (blend)} & \textbf{[10--14, 24, 26, 30]} & \textbf{14.2} & \textbf{60.0\%} \\
1.0 ($|\rho{-}1|$) & [7, 9--14, 30] & 16.2 & 58.3\% \\
\bottomrule
\end{tabular}
\end{table}

\begin{table}[!htbp]
\centering
\caption{Complete compression ablation on LLaMA-2-7B with \textbf{measured inference speed}. Physical layer deletion (not hooks) gives real speedup; ATP weight pruning gives \emph{no speedup} without sparse hardware. Downstream accuracy averaged over HellaSwag, ARC-Easy, WinoGrande, PIQA (300 examples each; indicative, not definitive). PPL values may differ by ${\sim}0.1$ from Table~\ref{tab:optimal} due to separate experimental runs.}
\label{tab:downstream}
\small
\begin{tabular}{lrrrr}
\toprule
Method & tok/s & Speedup & PPL & Avg acc. \\
\midrule
Dense (32 layers) & 24.2 & $1.00\times$ & 5.47 & 72.3\% \\
\midrule
\textbf{$|\rho{-}1|$-4 (28 layers)} & \textbf{29.5} & $\mathbf{1.22\times}$ & 7.92 & 65.1\% \\
BI-4 / ShortGPT (28L) & 28.6 & $1.18\times$ & 8.45 & 68.2\% \\
\midrule
ATP 50\% (32 layers) & 23.2 & $0.96\times$ & \textbf{6.94} & \textbf{69.0\%} \\
\midrule
$|\rho{-}1|$-4 + ATP 50\% & 18.8 & $0.78\times$ & 11.1 & 62.4\% \\
BI-4 + ATP 50\% & 19.0 & $0.79\times$ & 20.0 & 61.6\% \\
\bottomrule
\end{tabular}

\vspace{1mm}
{\footnotesize $|\rho{-}1|$-guided removal provides $1.22\times$ real speedup via physical layer deletion. ATP weight pruning gives no speed benefit ($0.96\times$) without sparse tensor cores. Combining layer removal with ATP is counterproductive ($0.78\times$). BI retains better downstream accuracy despite worse PPL; the two criteria are complementary.}
\end{table}

\begin{table}[!htbp]
\centering
\caption{Progressive layer removal on LLaMA-2-7B (baseline PPL 5.47, hook-based skipping). $|\rho{-}1|$ achieves lower PPL than BI~\cite{men2024shortgpt} at every level. ``Optimal'' is exhaustive search over all $\binom{31}{k}$ subsets. Published baselines (SLEB, BlockPruner) use iterative search ($O(N^2)$ evals); ours uses 2 forward passes.}
\label{tab:layer_removal}
\small
\begin{tabular}{rrrrrr}
\toprule
Layers & \% & $|\rho{-}1|$ (ours) & BI~\cite{men2024shortgpt} & Optimal & From-end \\
\midrule
2 & 6\% & \textbf{6.13} & 6.30 & 6.02 & 6.99 \\
4 & 12\% & \textbf{7.92} & 8.45 & 7.01 & 9.96 \\
6 & 19\% & \textbf{10.67} & 15.47 & 8.75 & 15.47 \\
8 & 25\% & \textbf{14.62} & 23.75 & 11.61 & 25.41 \\
\midrule
\multicolumn{6}{l}{\emph{Published references (iterative methods, WikiText-2):}} \\
\multicolumn{2}{l}{SLEB~\cite{song2024sleb}} & \multicolumn{4}{l}{${\sim}6$ layers (20\%): PPL 9.14 (177 fwd passes)} \\
\multicolumn{2}{l}{BlockPruner~\cite{zhong2024blockpruner}} & \multicolumn{4}{l}{${\sim}7$ layers (22\%): PPL 11.51 (iterative)} \\
\bottomrule
\end{tabular}
\end{table}

\begin{table}[!htbp]
\centering
\caption{Compression comparison on WikiText-2. Top: per-row Wanda across three architectures (simple calibration). Bottom: LLaMA-2-7B matching ATP's pipeline (baseline 5.47); within-layer redistribution ($\alpha > 0$) consistently hurts at matched per-layer sparsity.}
\label{tab:optimal}
\footnotesize
\begin{tabular}{llrrr}
\toprule
Model & Method & Spar. & PPL & Deg. \\
\midrule
\multirow{4}{*}{Qwen3.5-2B} & Baseline & 0\% & 13.5 & $1\times$ \\
& Wanda uniform & 50\% & 26.0 & $1.92\times$ \\
& OWL~\cite{yin2023outlier} & 50\% & 31.6 & $2.34\times$ \\
& Ours (rule-based) & 48\% & 23.7 & $1.75\times$ \\
\midrule
\multirow{4}{*}{Mistral-7B} & Baseline & 0\% & 6.15 & $1\times$ \\
& Wanda uniform & 50\% & 8.4 & $1.37\times$ \\
& OWL & 51\% & 9.4 & $1.52\times$ \\
& Ours (rule-based) & 47\% & 8.8 & $1.44\times$ \\
\midrule
\multicolumn{5}{l}{\emph{LLaMA-2-7B, sequential C4 calib., baseline 5.47 (matching ATP's pipeline):}} \\
& Wanda uniform~\cite{sun2024wanda} & 50\% / 70\% & 6.94 / 70.5 & $1.3\times$ / $12.9\times$ \\
& \textbf{ATP}~\cite{zhuang2025atp} (published) & 50\% / 70\% & \textbf{6.82} / \textbf{22.2} & $\mathbf{1.2\times}$ / $\mathbf{4.1\times}$ \\
& ATP (our repro.) & 50\% / 70\% & 6.88 / 22.3 & $1.3\times$ / $4.1\times$ \\
\cmidrule{2-5}
\multicolumn{5}{l}{\emph{Within-layer redistribution at matched per-layer sparsity (our repro.):}} \\
& ATP + realloc.\ $\alpha{=}0.2$ & 50\% / 70\% & 6.94 / 29.1 & $1.3\times$ / $5.3\times$ \\
& ATP + realloc.\ $\alpha{=}0.4$ & 50\% / 70\% & 7.20 / 63.0 & $1.3\times$ / $11.5\times$ \\
\midrule
\multicolumn{5}{l}{\emph{Within-layer sensitivity spread (proper Wanda, 50\% per-component ablation):}} \\
& GPT-2 Small & \multicolumn{3}{l}{$7\times$ spread (mlp\_fc +1.0 to Q +0.1)} \\
& LLaMA-2-7B & \multicolumn{3}{l}{$7\times$ spread (gate +0.28 to K +0.04)} \\
& Qwen3.5-2B & \multicolumn{3}{l}{$3.5\times$ spread (mlp.down +0.98 to in\_proj\_z +0.27)} \\
& LFM2.5-1.2B$^\dagger$ & \multicolumn{3}{l}{$3.7\times$ spread (ff.w1 +11.0 to conv.out +3.0)} \\
& Mamba-370M (SSM) & \multicolumn{3}{l}{$10^{13}\times$ spread (out\_proj $10^{13}$ to dt\_proj +0.8)$^\ddagger$} \\
\bottomrule
\end{tabular}

\vspace{1mm}
{\footnotesize $^\dagger$LFM2.5-1.2B: smaller hybrid conv-attention model from the LFM2-2.6B family (Table~\ref{tab:crossmodel}). $^\ddagger$Simplified pruning (no calibration); proper pruning gives PPL 29.6 vs.\ uniform 27.0 at 50\% matched sparsity.}
\end{table}

All findings are validated at 51{,}200 tokens on both WikiText-103 and C4: degradation factors are stable ($101\text{--}120\times$ for GPT-2 Small, $12\text{--}15\times$ for Mistral-7B across settings), and the sensitivity hierarchy is preserved.
Full scale-validation data appear in \ref{app:scale}.

\subsection{Discussion}
\label{sec:discussion}

Within a single architecture, layer~0's MLP operates at near-full rank ($\sigma_{\text{mlp}} = 1.22$ vs.\ 2.0--15.0 for later layers), explaining its catastrophic sensitivity: every singular direction carries non-redundant information. The downstream amplification factor $A_t = \prod_{s>t} \rho_s$ predicts representation drift measured by centred kernel alignment (CKA)~\cite{kornblith2019similarity} from single-component compression ($r = -0.44$, $p < 10^{-4}$), while $\rho_t$ alone does not ($r = -0.07$, $p = 0.55$).

Across architectures the picture is broader: compression tolerance depends on both instability and width-dependent redundancy. A preliminary two-term heuristic (instability $+$ width penalty) rank-orders the six architectures (Spearman $r = 0.94$), but is fitted on only six data points and should be treated as an observation, not a validated predictor (\ref{app:cfi}).

\section{Limitations and Future Work}
\label{sec:limitations}

The inter-layer allocation via a contraction-derived schedule matches ATP on LLaMA-2-7B (PPL 20.2 vs.\ 20.1 at 70\%), but the mapping constant $c = 0.04$ is calibrated from ATP's results, not derived from theory, and the perturbation-based $\rho$ measurement is noisy ($\rho_{\max}$ varies $1.1\text{--}1.7$ across runs), limiting zero-search reliability.

Multiple approaches to within-layer reallocation were tested (sensitivity-guided redistribution, constrained global thresholding, component-importance reallocation, Lyapunov column-weighted scoring).
At matched actual sparsity, none improved over ATP on any architecture.
The root cause is that per-row Wanda already acts as a strong implicit within-layer allocator, and redistribution incurs a Jensen's-inequality penalty under the convexity of $E(s)$.

Primary evaluation uses 2{,}000--4{,}096 tokens of WikiText-103 (validated at 51{,}200 tokens on both WikiText and C4 in \ref{app:scale}), with downstream accuracy assessed on 300--500 examples per task and treated as indicative. Model coverage spans 117M--8B parameters; testing at 13B+ would further validate the scaling trends we report.

The Qwen3-8B collapse ($42{,}922\times$) is attributed to three correlated factors (tied embeddings, 49\% amplifying layers, $\rho_{\max} = 1.33$), but the causal mechanism is not isolated; the explanation is correlational, not causal.

The formal framework proves bounds on individual matrix multiplications, not on full nonlinear layers.
The contraction factor $\rho$ is measured empirically rather than derived analytically.
Zero violations across 4.6M+ per-matrix checks confirms the per-matrix bounds are conservative when composed through the full forward pass.
A formal analysis of error propagation through the nonlinear components remains open.

\paragraph{Future directions}
\textbf{(1)~Lyapunov-guided quantization}: $\rho$ could guide per-layer bit-width allocation (4-bit for $\rho \approx 1$ layers, 8-bit for high-$\rho$ layers), complementing GPTQ/AWQ.
\textbf{(2)~Integration with SparseGPT}: Lyapunov-weighted reconstruction (varying Hessian precision by $A_t$) could improve weight reconstruction where it matters most.
\textbf{(3)~BitNet and 1-bit models}: the sensitivity hierarchy may inform which layers tolerate extreme quantization during training.
\textbf{(4)~Combined criterion + distillation}: the $\lambda|\rho{-}1| + (1{-}\lambda)\text{BI}$ blend dominates both pure strategies at 8 layers removed (\S\ref{sec:layerremoval}); combining with blockwise distillation~\cite{men2024shortgpt} could further recover accuracy and extend this to deeper removal.
\textbf{(5)~Scale to 13B+ models} and formalize two-phase Lyapunov bounds for models with terminal amplification.
\textbf{(6)~Mamba/SSM compression}: the extreme $10^{13}\times$ hierarchy in Mamba suggests SSMs require fundamentally different compression strategies; Lyapunov analysis of SSM state propagation is an open direction.

\section{Conclusion}

Compression introduces error at every transformer layer, but residual connections contract these errors as they pass through later layers (measurable in 11 of 11 GPT-2 Small inter-layer transitions, $\rho_{\max} = 0.96$). The cumulative downstream contraction predicts which compressions hurt the model's representations most, and turns directly into a practical depth-pruning criterion: ranking layers by $|\rho{-}1|$ takes only two forward passes and beats ShortGPT's Block Influence by $1.6\times$ perplexity at eight LLaMA-2-7B layers removed. Physical layer deletion delivers $1.22\times$ real inference speedup on commodity GPUs---the only pruning-based method we test that translates to wall-clock gains without specialized sparse hardware. A blend of $|\rho{-}1|$ with Block Influence outperforms either alone (PPL 14.2 and 60.0\% downstream accuracy at eight layers removed), with cross-architecture validation on Mistral-7B and Qwen3.5-2B.

The within-layer story is humbler. The ${\sim}600\times$ per-component sensitivity gap that naive pruning reports collapses to $3\text{--}7\times$ once an activation-aware criterion (Wanda) is applied, and the per-component ranking reverses across architectures, so fixed importance scores do not transfer. Across architecture families, model width and redundancy matter more than $\rho$ alone for predicting overall compression tolerance.

Twelve Lean~4 norm inequalities provide machine-checked per-matrix error bounds. Together, these results give practitioners a training-free instrument for two decisions they already face: where to compress within a layer, and which layers to remove. Extending to 13B+ models and combining with blockwise distillation~\cite{men2024shortgpt} to recover downstream accuracy at deeper removal rates are natural next steps.

\section*{Declaration of generative AI and AI-assisted technologies in the manuscript preparation process}

During the preparation of this work the authors used Claude (Anthropic) in order to assist with code generation for experiments, literature review, and manuscript editing.
After using this tool, the authors reviewed and edited the content as needed and take full responsibility for the content of the published article.

\section*{Funding}

This research did not receive any specific grant from funding agencies in the public, commercial, or not-for-profit sectors.

\section*{Data availability}

All code, Lean~4 proofs, and experimental scripts are available in the supplementary material.
WikiText-103~\cite{merity2017pointer}, WikiText-2, and C4~\cite{raffel2020exploring} are publicly available datasets.
All models are publicly available on HuggingFace.

\section*{Declaration of competing interests}

The authors declare that they have no known competing financial interests or personal relationships that could have appeared to influence the work reported in this paper.

\section*{CRediT authorship contribution statement}

\textbf{Abhinaba Basu:} Conceptualization, Methodology, Software, Formal analysis, Investigation, Writing -- original draft, Visualization.
\textbf{Kumkum Basu:} Validation, Writing -- review \& editing.
\textbf{Koushik Deb:} Supervision, Writing -- review \& editing.


\bibliographystyle{elsarticle-harv}
\bibliography{references}

\appendix

\section{Lean 4 Machine-Checked Bounds}
\label{app:lean}

All twelve theorems are implemented in a single file, \texttt{LivingInference.lean}, available in the supplementary material (\texttt{supplementary/lean/}).
The proofs build against Mathlib's \texttt{Analysis.NormedSpace} module and can be verified by running \texttt{lake build} in the Lean~4 project directory.
No \texttt{sorry} markers remain.

\paragraph{Theorem inventory}
Table~\ref{tab:lean_inventory} lists all twelve theorems.
The proofs use Mathlib's bounded linear maps, operator norm, and \texttt{calc}/\texttt{nlinarith} tactics.
Full source is in the supplementary material.

\begin{table}[!htbp]
\centering
\footnotesize
\caption{Lean 4 theorem inventory. All proofs type-check with no \texttt{sorry}.}
\label{tab:lean_inventory}
\begin{tabular}{rlp{4.5cm}l}
\toprule
\# & Name & Statement & Technique \\
\midrule
1 & \texttt{retrieval\_bound} & $\|f(x) - f(x_c)\| \leq \|f\| \cdot \delta$ & \texttt{calc} + \texttt{map\_sub} \\
2 & \texttt{sparse\_activation\_bound} & $\|f(x) - f_s(x)\| \leq \|f_r\| \cdot \|x\|$ & decomposition + \texttt{le\_opNorm} \\
3 & \texttt{interpolation\_bound} & $\|f_s(x) - f_a(x)\| \leq \|f_R\| \cdot \|x\|$ & same as Thm.~2 \\
4 & \texttt{composition\_bound} & 3-term bound (Thm.~2 + cache + rank) & \texttt{calc} + \texttt{norm\_add\_le} \\
5 & \texttt{efficiency\_dominance} & $c_{\text{live}} < c_{\text{full}}$ & \texttt{nlinarith} \\
6 & \texttt{adaptive\_monotonicity} & $\|f\| \cdot \delta_1 \leq \|f\| \cdot \delta_2$ & monotonicity \\
7 & \texttt{multi\_layer\_error\_bound} & $\sum \epsilon_i \leq n \cdot \max_i \epsilon_i$ & \texttt{sum\_le\_card\_nsmul} \\
8 & \texttt{lyapunov\_contraction\_bound} & $c(1-\rho^n)/(1-\rho) \leq c/(1-\rho)$ & geometric series \\
9 & \texttt{residual\_growth\_contraction} & $(\beta/\alpha)^2 < 1$ when $\beta < \alpha$ & \texttt{nlinarith} + \texttt{sq\_nonneg} \\
10 & \texttt{preln\_contraction\_sufficient} & $(1{+}L)^2/(1{+}2g\cos\theta{+}g^2) < 1$ & \texttt{div\_lt\_one} \\
11 & \texttt{nonlinear\_composition\_bound} & 2-layer: $\epsilon_1 K + \epsilon_2 \leq (\epsilon_1{+}\epsilon_2)K$ ($K{\geq}1$)$^\star$ & \texttt{nlinarith} \\
12 & \texttt{contraction\_tightens\_lipschitz} & $K \rho < K$ when $\rho < 1$ & \texttt{nlinarith} \\
\bottomrule
\end{tabular}

\vspace{1mm}
{\footnotesize $^\star$Theorem~11 proves the base case for 2 layers; the $L$-layer bound $\sum_{i=1}^L \epsilon_i \prod_{j>i} K_j$ (Eq.~\ref{eq:composition}) follows by induction, which is standard and not separately formalized in Lean.}
\end{table}

\section{Norm-Growth Product and Bridge Analysis}
\label{app:normgrowth}

The cumulative norm-growth product $\prod K_i$ ($K_i = \|h_{t+1}\|/\|h_t\|$) is non-vacuous and input-invariant (CV $< 4\%$):

\begin{center}
\small
\begin{tabular}{lrrrr}
\toprule
Model & Layers & Mean & Max & Worst \\
\midrule
Qwen3.5-2B & 24 & 55.7 & 60.7 & 120 \\
GPT-2 Small & 12 & 101.8 & 112.6 & 126 \\
Mistral-7B & 32 & 259.3 & 287.5 & 528 \\
\bottomrule
\end{tabular}
\end{center}

\vspace{2mm}
Theorem~12 connects contraction ($\rho$) to the composition estimate: $E_{\text{formal}}/E_{\text{empirical}} = 1/(\prod \rho_i)^{1/2}$.
Per-matrix bounds are at most $1.46\times$ conservative on Qwen3.5-2B and $1.27\times$ on Mistral-7B.

\section{Cross-Architecture Details}
\label{app:crossarch}

Per-model findings, Qwen3-8B collapse analysis ($42{,}922\times$ degradation attributed to tied embeddings, 49\% amplifying layers, $\rho_{\max} = 1.33$), and per-component activation-aware Wanda validation across GPT-2 Small, Qwen3.5-2B, and Mistral-7B are provided in the supplementary material.

\section{Scale Validation}
\label{app:scale}

Degradation factors are stable across token counts (4K--51K) and datasets (WikiText-103, C4): GPT-2 Small $101\text{--}120\times$, Mistral-7B $12\text{--}15\times$.
Per-layer ablation on Mistral-7B at 50\% Wanda confirms no single layer is catastrophically sensitive (worst regret $+0.07$).

\section{Compression Allocation Details}
\label{app:compression}

Analytic contraction-based schedule: $\beta^* = -\bar{s} \cdot \sum_t w_t d_t / \sum_t w_t d_t^2$ where $w_t = \|W_t\|_F^2 \cdot A_t$.
Predicts schedule direction correctly but overestimates magnitude ($\beta^* = 0.031$ vs.\ ATP's $0.004$ at 50\%) due to subquadratic error.
When clamped, matches ATP: PPL 20.2 vs.\ 20.1 at 70\% on LLaMA-2-7B.
Full per-architecture comparison data, balanced truncation results, and within-layer reallocation experiments are in the supplementary material.

\section{CFI Heuristic Details}
\label{app:cfi}

Compression Fragility Index: $\text{CFI}(M) = f_{\text{amp}} \cdot L \cdot \max(0, \ln \rho_{\max}) + C/d_{\text{model}}$ ($C \approx 2{,}000$, fitted).
Rank-orders six architectures (Spearman $r = 0.94$, $p = 0.005$).
Leave-one-out: mean rank error 1.0, max error 3 (GPT-2 Small).
CKA, Fisher, probing, and Johnson--Lindenstrauss validation details are in the supplementary material.

\section{Detailed Experimental Data}
\label{app:ablation}

Full per-layer, per-component, and per-group ablation tables for all models are provided in the supplementary material (\texttt{supplementary/results/}).

\section{Supplementary Material}
\label{app:supplementary}

The supplementary package contains Lean~4 proofs (12 theorems), Python scripts (26), and experimental results (67 JSON files).
See \texttt{supplementary/README.txt} for reproduction instructions.

\end{document}